\documentclass[conference]{IEEEtran}
\IEEEoverridecommandlockouts
\usepackage{cite}
\usepackage{amsmath,amssymb,amsfonts}

\usepackage[utf8]{inputenc} 
\usepackage[T1]{fontenc}    
\usepackage{hyperref}       
\usepackage{url}            
\usepackage{booktabs}       
\usepackage{amsfonts}       
\usepackage{amsthm}
\usepackage{nicefrac}       
\usepackage{microtype}      
\usepackage{xcolor}         
\usepackage{algorithm} 
\usepackage{graphicx}
\usepackage{algpseudocode}
\usepackage{derivative}
\usepackage{amsmath}

\newtheorem{prob}{Problem Definition}
\def\BibTeX{{\rm B\kern-.05em{\sc i\kern-.025em b}\kern-.08em
    T\kern-.1667em\lower.7ex\hbox{E}\kern-.125emX}}
    
\begin{document}
\title{Detection of Unknown-Unknowns in Human-in-Loop Human-in-Plant Systems Using Physics Guided Process Models$^*$\thanks{$^*$This work is funded in part by the DARPA AMP grant (N6600120C4020). Opinions in the paper are those of the authors' and not endorsed by the agency.}
}
\author{\IEEEauthorblockN{Aranyak Maity}
\IEEEauthorblockA{\textit{Impact Lab, Arizona State University} \\
Tempe, Arizona, USA \\
amaity1@asu.edu}
\and
\IEEEauthorblockN{Ayan Banerjee}
\IEEEauthorblockA{\textit{Impact Lab, Arizona State University} \\
Tempe, Arizona, USA \\
abanerj3@asu.edu}
\and
\IEEEauthorblockN{Sandeep K.S. Gupta}
\IEEEauthorblockA{\textit{Impact Lab, Arizona State University} \\
Tempe, Arizona, USA \\
sandeep.gupta@asu.edu}
}

\maketitle

\begin{abstract}
Unknown-unknowns are operational scenarios in artificial intelligence (AI) enabled autonomous systems (AAS) that are not accounted for in the design and validation phase. Their occurrences can be attributed to uncertain human interaction and rigors of regular usage in deployment. The paper illustrates that integrated modeling of human user and AAS from the human-in-the-loop human-in-the-plant (HIL-HIP) perspective can potentially enable detection of unknown unknowns before it results in fatal safety violations. We propose a novel framework for analyzing the operational output characteristics of safety-critical HIL-HIP systems to discover unknown-unknown scenarios and evaluate potential safety hazards. We propose dynamics-induced hybrid recurrent neural networks (DiH-RNN) to mine a physics-guided surrogate model (PGSM) that checks for deviation of the AAS from safety-certified operational characteristics. The PGSM enables early detection of unknown-unknowns based on the physical laws governing the system. We demonstrate the detection of operational changes in an Artificial Pancreas(AP) due to unknown insulin cartridge errors.
\end{abstract}


\section{Introduction}
Safety critical artificial intelligence (AI) enabled autonomous system (AAS) such as Artificial Pancreas (AP)~\cite{weaver2018hybrid} and Semi-Autonomous Driving (SAD)~\cite{SynthesisSeshia} operate in a human-in-the-loop (HIL) paradigm, where the AAS operates with full autonomy under most input conditions while requiring manual override for potential catastrophic safety loss~\cite{SynthesisSeshia}. Diversity in human interaction driven by their social-behavioral-economic background~\cite{banerjee2021socio} and rigorous usage with human-in-the-plant (HIP), can result in novel scenarios or previously unknown component failures. These un-tested scenarios can result in unknown unknowns (U2s) in safety-certified AAS with potentially fatal consequences. This paper enables early detection of U2s before safety violations. 

\begin{figure}
\center
\includegraphics[trim=0 0 0 0,width=0.9\columnwidth]{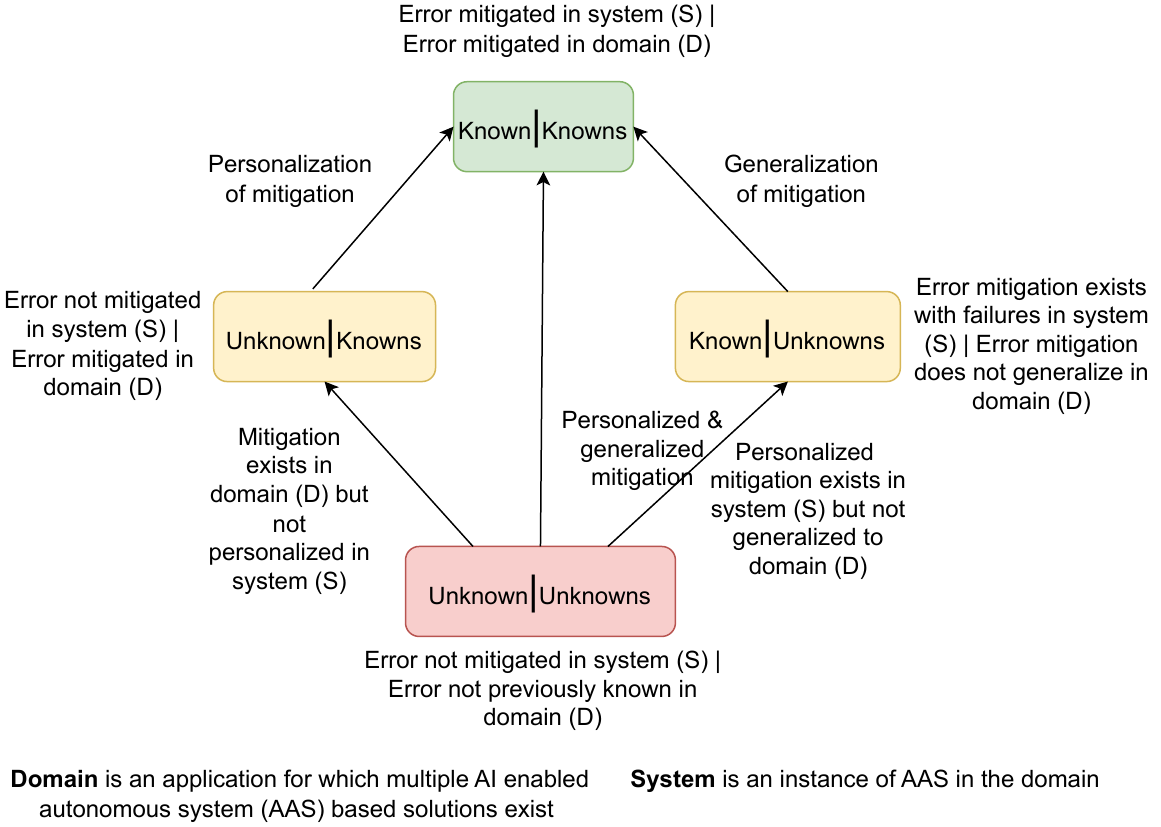} 
 \caption{Error lattice categorizing different types of errors based on how they are handled in safety certification and operational deployment of AAS.}
 \vspace{-0.2 in}
    \label{fig:lattice}
\end{figure}

\subsection{Unknown Unknowns}
Any safety critical AAS has to undergo safety certification before deployment in the application domain~\cite{Lamrani21Certification}. In this phase, the AAS is tested for known errors in the domain to check for safety hazards. From the certification perspective, an error lattice can be envisioned (Fig. \ref{fig:lattice}), where each node is a tuple $\{S | D\}$, indicating whether (or not) an error is tested for in the system $S$, known (unknown), and whether (or not) an error has been previously observed in the domain, known (unknown). 

If a targeted mitigation strategy exits in the AAS that avoids serious failures (even though it cannot eliminate the error) and also is common knowledge in the domain, then it is a (known | known ) error. Examples of (known | known)s include glucose sensor error in AP, and angle of attack sensor errors in an aircraft~\cite{banerjee2023statistical}. If the mitigation strategy has significant failure rates in the AAS as well as in other systems in the domain, then the error is (known | unknown). Examples include predictive hypoglycemia alerts in AP, or object detection in SAD. If a mitigation strategy is not present in AAS even though the error is previously known in the domain, then the error is a (unknown | known)s. Examples include safe maneuvering control augmentation system (MCAS) disengage module that was disabled for buyers who opt out~\cite{banerjee2023statistical}.

An (unknown | unknown) error has not been tested in the AAS certification process, and is also previously unknown in the domain. Examples include insulin cartridge error, before it actually occurred in Medtronic 670G AP. The AAS design can move from one node to the other through personalized mitigation and generalization to the domain (Fig. \ref{fig:lattice}).

\subsection{Drawbacks of controller design approaches}
 Broadly, two AAS design approaches exist: a) optimal control for satisfying both efficacy and safety goals using techniques such as supervised data-driven learning~\cite{dawson2022safe}, and b) human-in-the-loop (HIL) controller synthesis~\cite{SynthesisSeshia}. In the first approach, the learning system simultaneously learns control action and control Lyapunov barrier functions (CLBF), which act as a safety certificate~\cite{dawson2022safe}. This class of approaches considers the system as a control affine and human inputs as external perturbations within a distribution. While such approaches work well for full automation, in HIL safety critical AAS, human inputs in operational deployment bear a causal relationship with observations of both short and long-term safety and efficacy and hence may not be captured merely by a stochastic process~\cite{banerjee2023high}.
 
HIL controller synthesis approach has focused on optimal / model predictive / reinforcement learning-based control that incorporates a model of human action in the control design and decides on optimal switching between the AI controller and the human inputs~\cite{SynthesisSeshia}. The human user in safety-critical AAS is frequently advised by agents (endocrinologists for AP or the car manufacturers and national transportation safety board in SAD) about context-dependent personalized actions to optimize efficacy and avoid safety hazards. However, in HIL-HIP safety critical AAS \cite{lamrani2020operational}, the perception of impending safety hazards can cause the human user to apply such actions out of context. As such modeling human action is complex, resulting in un-modelled scenarios in the HIL controller synthesis approach. 
  \begin{figure}
\center
\includegraphics[trim=0 0 0 0,width=0.9\columnwidth]{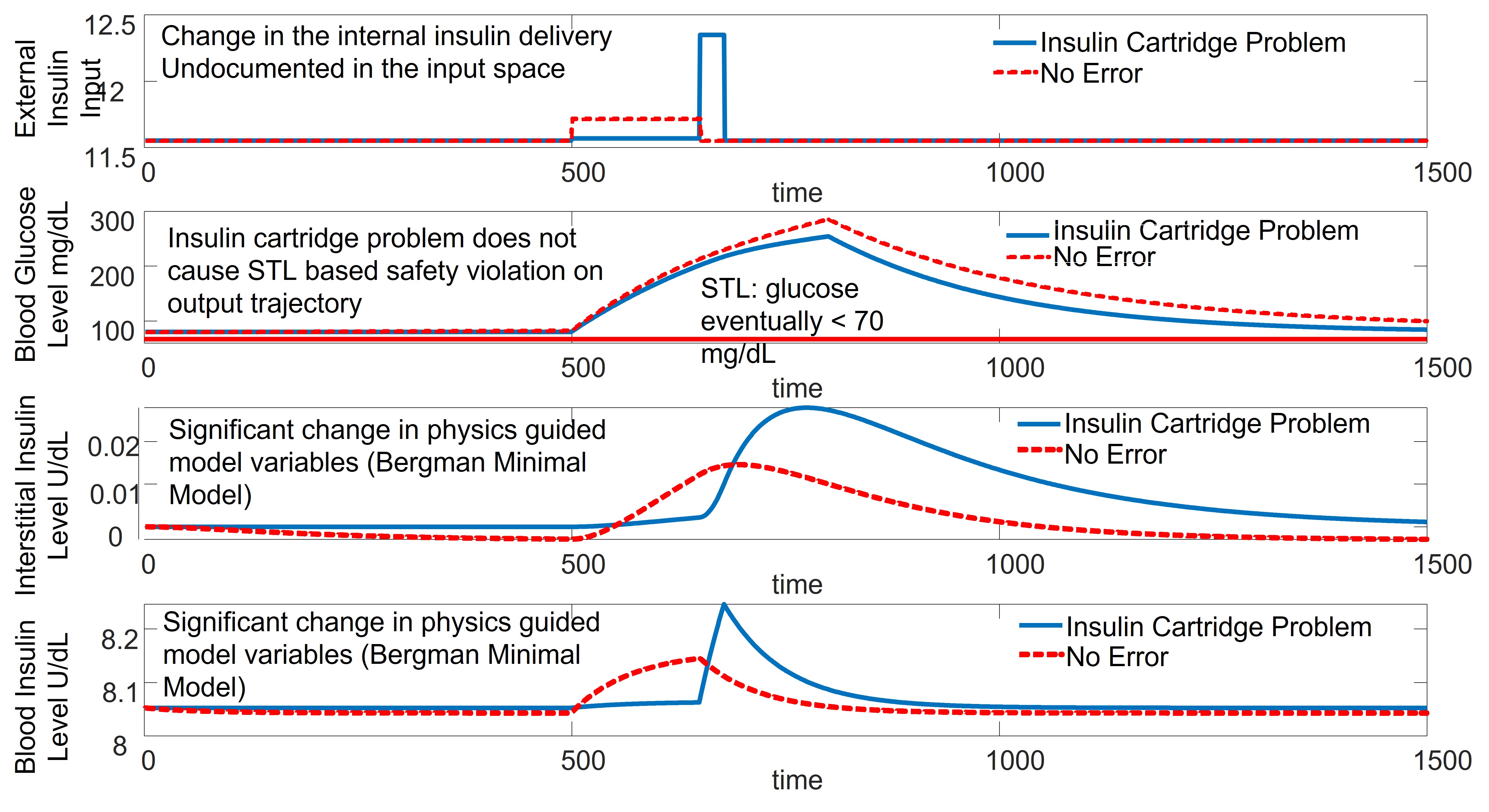} 
 \caption{Insulin Cartridge Problem in Medtronic 670G. The glucose level does not show any safety violation, although the operational model (solid line) has a significant deviation from the original model (dashed line). }
    \label{fig:APEx}
\end{figure} 
\begin{figure}
\center
\includegraphics[trim=0 0 0 0,width=0.75\columnwidth]{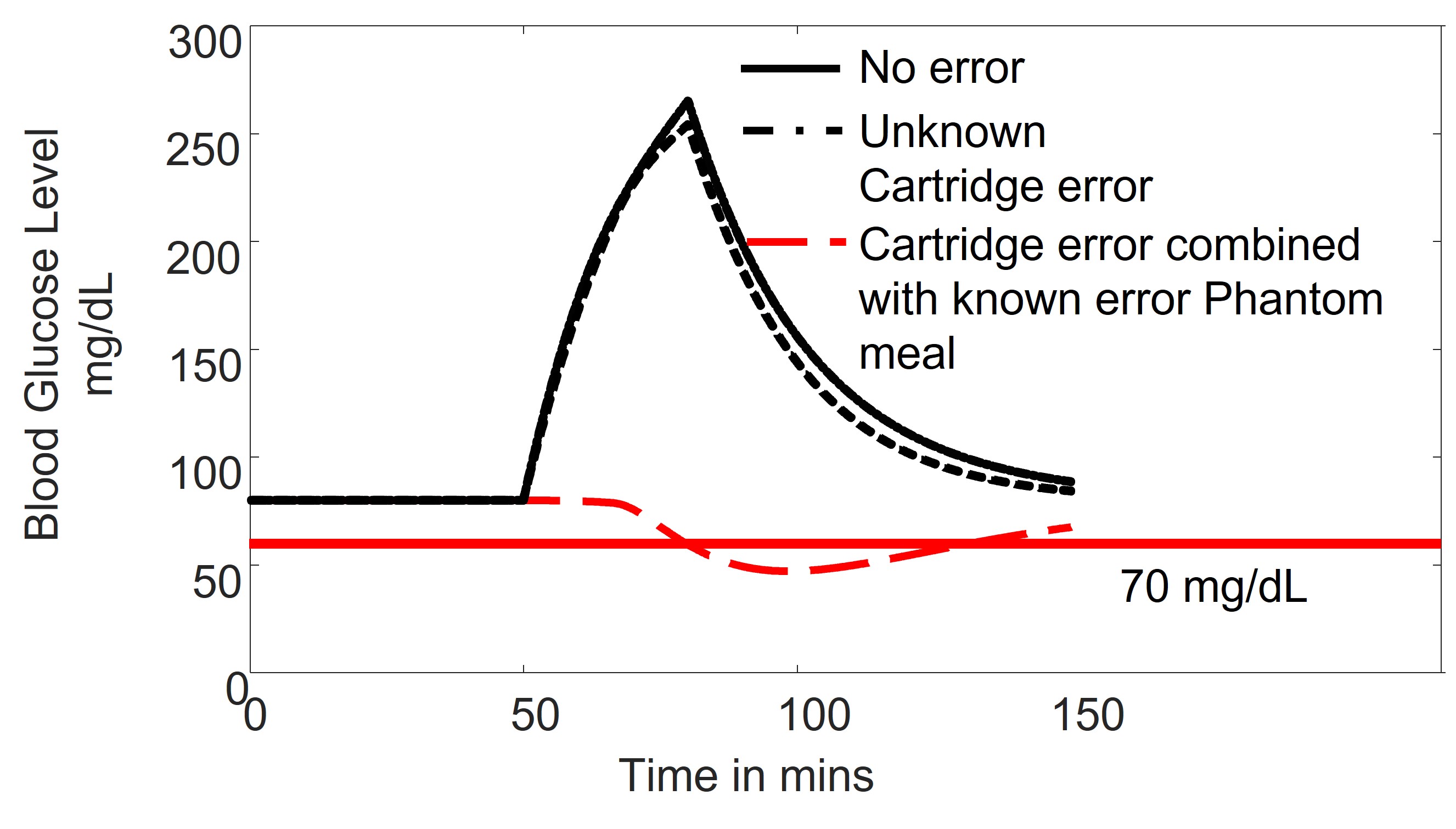} 
 \caption{Unknown insulin cartridge problem when combined with the known fault of phantom meal violates STL properties for the same input space. }
    \label{fig:APExV2}
\end{figure}
\subsection{Types of U2s and their aggregate effects}
In this paper, we restrict the U2s to encompass: a) un-modelled human actions, where the user of the AAS provides an input that is not certified safe during test time. b) latent sensor/ actuator errors, where a system component such as a sensor or an actuator fails or encounters faults that are previously unknown. As a result, despite significant advances in safety engineering, AAS often fails with fatal consequences. Some failures are \textit{unintentional}, highlighted in recent crash reports from Tesla~\cite{Tesla,banerjee2021faultex}, lawsuit on Medtronic for their automated insulin delivery (AID) system causing 1 death and 20,000 injuries~\cite{Medtronic}, and some are \textit{intentional}, Volkswagon cheating case~\cite{BiewerDoping,maity2022cyphytest}.
The fault signatures are also unknown and hence can be misconstrued by the user as an efficacy loss to be mitigated by the advisory agent-guided personalization action. However, in reality, such personalization action can combine with the latent fault signature to result in fatal hazards. Insulin cartridge error in AP (or AID) was one such error, that stunted insulin delivery with no signature in the pump logs. Fig. \ref{fig:APEx} shows an example of the Insulin cartridge error where the pump signature doesn't show much deviation.  The user would only perceive that the insulin delivery by the pump is not controlling a high glucose excursion. The user would follow the advisory agent-approved personalization action of administering a correction bolus either through the normal pump operational mechanism or by subverting it, a phenomenon called a phantom meal. In reality, the insulin stacking effect is induced, which eventually results in an unprecedented amount of insulin being delivered in a very short time, inducing severe hypoglycemia~\cite{Medtronic}. Fig. \ref{fig:APExV2} illustrates such a scenario.

\begin{figure}
\center
\includegraphics[trim=0 40 0 0,width=\columnwidth]{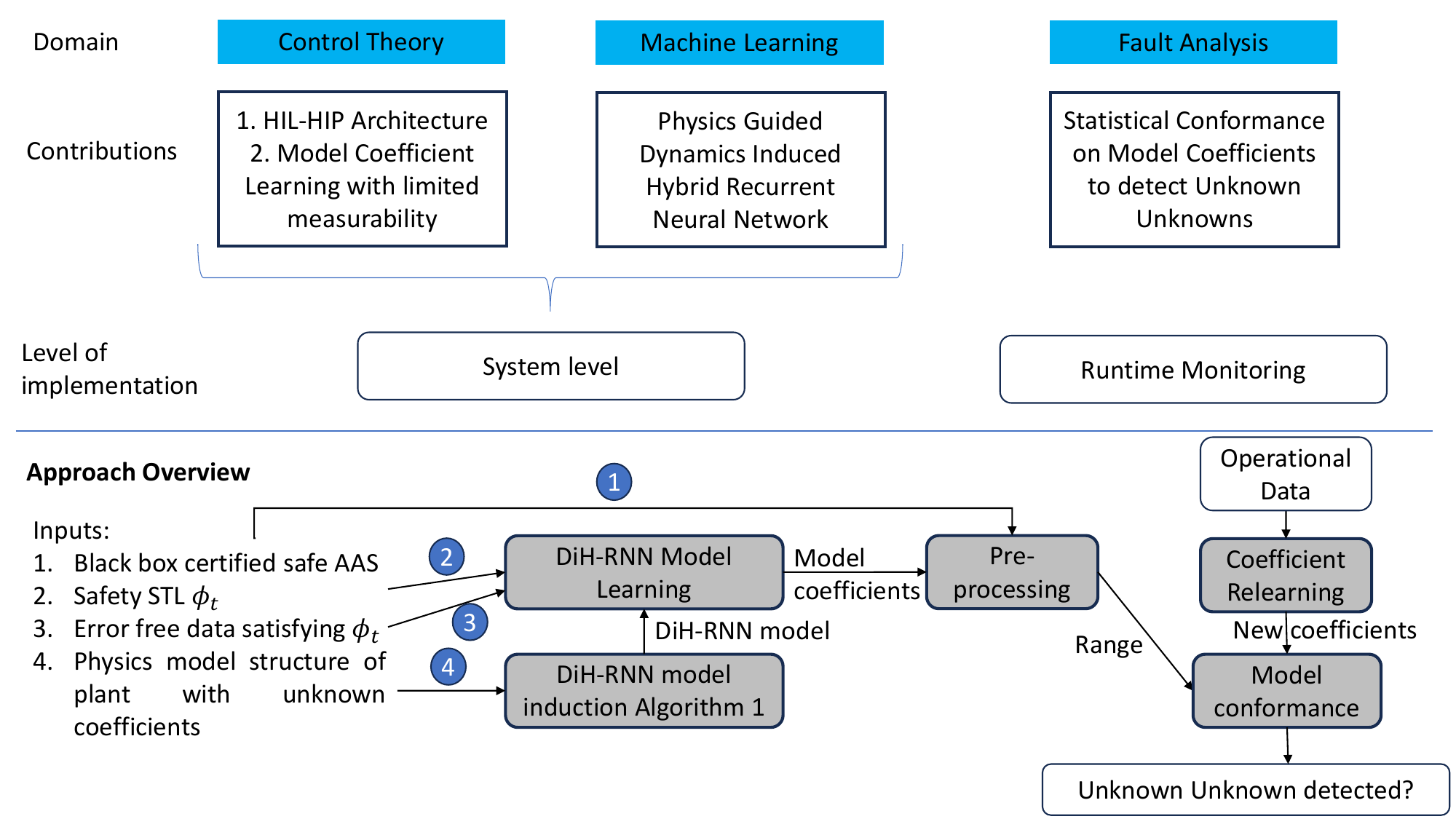} 
 \caption{Overview of Contributions and approach.}
    \label{fig:Approach}
\end{figure}

\subsection{Overview of Approach}
The central hypothesis of our approach is that in presence of unknown unknowns, even if AAS output demonstrates normal operation, the physical laws guiding the plant with HIP components that affects the input output relations will be violated. With this assumption, our approach (Fig. \ref{fig:Approach}) takes the following inputs: a)	Black box autonomous system tested to be safe in the lab, b) Safety condition expressed using signal temporal logic (STL), c) Error free data showing satisfaction of safety STL, d) Structure of physics model of the plant given by domain experts with unknown model coefficient values, and provides answer to the following question: ``Does the system have any unknown unknown errors?"

Our paper has three main contributions in different domains:

\noindent\textbf{a) Control system:} Integrated modeling the AAS as a HIL-HIP system and physics guided surrogate model (PGSM) coefficient value extraction from data that has unmeasurable system variables.

\noindent\textbf{b) Machine learning:} Novel neural network (NN) structure of Dynamics Induced Hybrid Recurrent NN (DiH-RNN), such that its back propagation results in a unique solution to the model coefficient value extraction problem.

\noindent\textbf{c) Fault Analysis:} Statistical model conformance on model coefficients to determine unknown unknowns.

\noindent\textbf{Unknown unknown detection} has the following steps: 

\textit{Pre-processing:} Using the MSI-RNN, AIIM first learns model coefficients for safety certified actions

\textit{Unknown Unknown detection:} a)	AIIM periodically relearns model coefficients, b)	Uses model conformance to check deviation, c)	If new model coefficients deviate from pre-processing step, raises alarms for unknown unknown errors.

\section{Preliminaries}
\textbf{Physics Model:} A physics model is a dynamical system expressed using a system of linear time-invariant ordinary differential equations in Eqn. \ref{eqn:set}. The system has $n$ variables $x_i$, $i \in \{1\ldots n\}$ in an $n \times 1$ vector $\mathcal{X}$, $\mathcal{A}$ is an $n \times n$ coefficient matrix, $\mathcal{B}$ is an $n \times n$ diagonal coefficient matrix.
\begin{equation}
\label{eqn:set}
\scriptsize
\dot{X(t)} = \mathcal{A}X(t) + \mathcal{B}U(t),\text{  } Y(t) = \beta X(t) 
\end{equation}
where $U(t)$ is a $n \times 1$ vector of external inputs. $Y(t)$ is the $n \times 1$ output vector of the system of equations. An $n \times n$ diagonal matrix, $\mathcal{\beta}$ of 1s and 0s, where $\beta_{ii} = 1$ indicates that the variable $x_i$ is an observable output else it is hidden and is not available for sensing.

A formal object $\hat{\mu}$ is a physics model when the set of models $\mu$ can be described using the coefficient $\omega =\mathcal{A} \bigcup \mathcal{B}$. The formal object can then take any $\theta$ as input and given the model coefficients $\omega$, generate a trace $\zeta_\theta = \hat{\mu}(\omega,\theta)$. 

\textbf{Trajectory and Models:} 
A trajectory $\zeta$ is a function from a set $[0, T]$ for some $T \in \mathcal{R}^{\geq 0}$ denoting time to a compact set of values $\in \mathcal{R}$. The value of a trajectory at time $t$ is denoted as $\zeta(t)$. Each trajectory is the output of a model $M$. A model $M$ is a function that maps a $k$ dimensional input $\theta$ from the input space $\Theta \subset \mathcal{R}^k$ to an output trajectory $\zeta_\theta$.     

The input $\theta \in \Theta$ is a random variable that follows a distribution $\mathcal{D}_\Theta$. The model $M$, can be simulated for input $\theta$ and a finite sequence of time $t_0 \ldots t_n$ with $n$ time steps and generate the trajectory $\zeta_\theta$ such that $\zeta_\theta(t_i) = \Sigma(\theta,t_i)$.

\textbf{Trace:}
Concatenation of $p$ output trajectories over time $\zeta_{\theta_1} \zeta_{\theta_2} \ldots \zeta_{\theta_p}$ is a trace $\mathcal{T}$.

\textbf{Continuous model mining:} 
Given a trace $\mathcal{T}$, continuous model mining maps the trace into a sequence $\Omega$ of $p$, $\omega_i$s such that $\forall i$  $dist(\hat{\mu}(\omega_i,\theta_i),\zeta_{\theta_i}) < \upsilon$, where $dist(.)$ is a distance metric between trajectories and $\upsilon \approx 0$ is decided by the user.

 \subsection{Signal Temporal Logic}
  Signal temporal logic are formulas defined over trace $\mathcal{T}$ of the form $f(\Omega) \geq c$ or $f(\Omega) \leq c$. Here $f: \mathcal{R}^p \rightarrow \mathcal{R}$ is a real-valued function and $c \in \mathcal{R}$. STL supports operations as shown in Eqn. \ref{eqn:STL}.
\begin{equation}
\label{eqn:STL}
\scriptsize
\phi, \psi := true | f(\Omega) \geq c | f(\Omega) \leq c | \neg \phi | \phi \wedge \psi | \phi \vee \psi|F_I \phi | G_I \phi | \phi U_I \psi,
\end{equation}  
   where $I$ is a time interval, and $F_I$, $G_I$, and $U_I$ are eventually, globally, and until operations and are used according to the standard definitions\cite{donze2010robust}\cite{fainekos2009robustness}.
To compute a degree of satisfaction of the STL we consider the robustness metric.
   The robustness value $\rho$ maps an STL $\phi$, the trajectory $\zeta$, and a time $t \in [0, T]$ to a real value. An example robustness $\rho$ for the STL $\phi: f(\Omega) \geq c$ is $\rho(f(\Omega)\geq c,\Omega,t) = f(\Omega(t)) - c$.

\subsection{Conformal Inference}
\label{sec:ModelConformance}
 \begin{figure}
\center
\includegraphics[trim=0 150 0 0,width=\columnwidth]{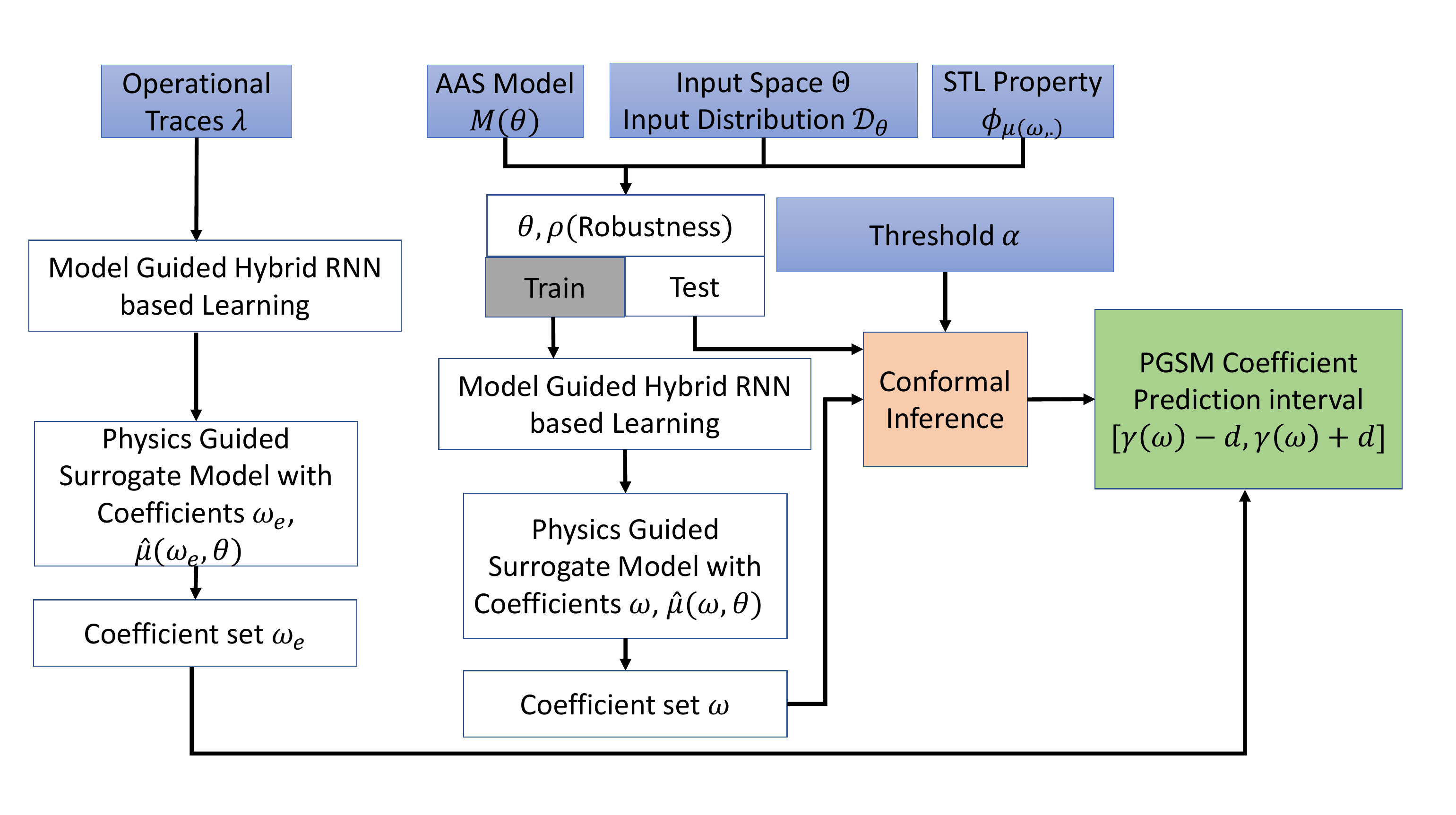} 
 \caption{Overview of the Conformal Inference Approach motivated by \cite{qinstatistical}.}
    \label{fig:OverviewApp}
\end{figure} 

In this approach, we use error free operational traces to learn the PGSM model coefficients $\omega_e$ (Fig. \ref{fig:OverviewApp}). The input space $\Theta$ s partitioned into train and test set. An STL $\phi_\mu(\omega,\omega_e)$ is defined as globally it is true that the largest normalized model coefficient deviation is less than 0.1. Consequently, its robustness is a quantification of the difference in model coefficient values. For each training and test data, we extract the model coefficients using DiH-RNN, and compute the mean robustness value $\rho_m$ across training data. From the test data $x_t$ we compute the mean robustness residue $\Tilde{\rho} = \rho_{x_t} - \rho_m$ and its variance $s(\Tilde{\rho})$. Any new data with unknown unknowns should result in model coefficients such that the STL robustness residue is beyond the range $[\Tilde{\rho}-s(\Tilde{\rho}),\Tilde{\rho}+s(\Tilde{\rho})]$.

\section{Learning a Physics-Driven Surrogate Model }
A surrogate model is a quantitative abstraction of the black box AAS model $M$. A quantitative abstraction satisfies a given property on the output trajectory of the model. In this paper, this quantitative property is the robustness value of an STL property. With this setting, we define a $\delta$-surrogate model $\hat{\mu}$.  \\
\textbf{($\delta$, $\epsilon$) Probabilistic Surrogate model:} Let $\zeta_\theta$ be a trajectory obtained by simulating $M$ with input $\theta$. Let $\omega^T$ be the coefficients of the physics-guided representation of the original model. Given a user-specified $\epsilon$, the formal object $\hat{\mu}(\omega,\theta)$ is a ($\delta$,$\epsilon$) probabilistic distance preserving surrogate model if 
\begin{equation}
\label{def:DE} 
\scriptsize
\exists \delta \in \mathcal{R}, \epsilon \in [0,1] : P(|\rho(\phi,\omega^T)-\rho(\phi,\omega)| \leq \delta) \geq 1-\epsilon.
\end{equation}
A $\delta$ surrogate model guarantees that the robustness value evaluated on a physics model coefficient $\omega$ derived from the trajectory $\zeta_\theta$ will not be more than $\delta$ away from the robustness computed on the coefficients of the original AAS model $M$.

\section{Coefficient mining from trajectory}
\begin{prob}
\label{prob:Prob}
Given a set of variables $\mathcal{X}(t)$, a set of inputs $U(t)$, a $\beta$ vector indicating observability, and a set $\mathcal{T}$ of traces such that $\forall i : \beta_i = 1 \exists T(x_i) \in \mathcal{T} $ and $\forall u_j(t) \in U(t) \exists T(u_j) \in \mathcal{T}$.

\noindent{\bf Derive:} approximate coefficients $\mathcal{A}^\prime$ and $\mathcal{B}^\prime$ such that: 
\begin{itemize}
\item $\forall i,j$ $|\mathcal{A}^\prime (i,j)-\mathcal{A}(i,j)| < \xi$
\item $\forall i$ $|\mathcal{B}^\prime(i,i)-\mathcal{B}(i,i)| < \xi$
\item Let $\mathcal{T}^a$ be the set of traces that include variables derived from the solution to differential equation $\frac{dX(t)}{dt} = \mathcal{A}^\prime X(t) + \mathcal{B}^\prime U(t)$ then $\forall i : \theta_i = 1,$ and $\forall k \in \{1 \ldots N\}, |T^a(x_i)[k] - T(x_i)[k]| < \Psi T(x_i)[k]$,
\end{itemize}   
where $\xi$ is the error in the coefficient estimator, while $\Psi$ is the error factor for replicating the traces of variables with the estimated coefficients. 
\end{prob}
\subsection{Dynamics Induced RNN}
For each variable $x_i \in X$ the system of dynamical equations takes the form of Eqn. \ref{eqn:ind}.
\begin{equation}
\label{eqn:ind}
\scriptsize
\dot{x_i} = \sum_{j = 1}^{n}{a_{ij} x_j} + b_{ii}u_i.
\end{equation}
\begin{algorithm}
	\caption{RNN induction algorithm}
 \scriptsize
\begin{algorithmic}[1]
		\State $\forall x_i \in X$ create an RNN cell with $n + 1$ inputs and  $x_i$ as the hidden output.   
		\For {each RNN cell corresponding to $x_i$}
		 	\For {each $j \in {1 \ldots n}$}  
				\If {$a_{ij} \neq 0$}
					\State Add connection from the output of $x_j$ RNN cell to the input of $x_i$ RNN cell.				
				\EndIf
			\EndFor	
			\State Remove all other inputs in the RNN which does not have any connection.
			
			\For {each $j \in {1 \ldots n}$}  
				\If {$b_{ij} \neq 0$}
					\State Add $u_j$ as an external input to the RNN cell for $x_i$.				
				\EndIf
			\EndFor				
			
		\EndFor
		\State Assign arbitrary weights to each link.
\end{algorithmic}
	\label{alg:Induction}
\end{algorithm}
We explain Algorithm \ref{alg:Induction} using the linearized Bergman Minimal Model (BMM) as an example. The model is a dynamical system that mimics the glucose-insulin biochemical dynamics in the human body. The Bergman Minimal model is linearized using Taylor Series expansion starting from overnight glucose dynamics and going up to time $N$. The linearized model is represented in Eqn. \ref{eqn:5}, \ref{eqn:6} and \ref{eqn:7}.
\begin{equation}
\label{eqn:5}
\scriptsize
{\dot{\Delta i}(t)} = -n\Delta i(t) + p_4 u_1(t)
\end{equation}
\begin{equation}
\label{eqn:6}
\scriptsize
 {\dot{\Delta i_s}(t)} = -p_1 \Delta i_s(t) + p_2 (\Delta i(t) - i_b)
 \end{equation}
 \begin{equation}
 \scriptsize
\label{eqn:7}
{ \dot{\Delta G}(t)} = -\Delta i_s (t) G_b -p3 (\Delta G(t)) + u2(t)/VoI,
\end{equation}
The symbol $\Delta$ means difference from the initial condition, e.g., $\Delta G = G(t) - G(0)$. The input vector $U(t)$ consists of the overnight basal insulin level $i_1b$ and the glucose appearance rate in the body $u_2$. The output vector $Y(t)$ comprises the blood insulin level $i$, the interstitial insulin level $i_s$, and the blood glucose level $G$. For this example, we consider that only the blood glucose level $G$ is a measurable output of the system. $i_s$ and $i$ are intermediate outputs that are not measurable for the system of equations and only contribute to the final glucose output. $p_1$, $p_2$, $p_3$, $p_4$, $n$, and $1/V_o I$ are all the coefficients of the set of differential equations. Algorithm \ref{alg:Induction} gives the AP DiH-RNN.
\subsection{Forward pass in DiH-RNN}
We prove that the Forward pass on an RNN cell estimates the solution of Eqn. \ref{eqn:set} with error factor $\Psi$ if $\tau \leq min_{i}{\frac{\sqrt{2\Psi}}{a_{ii}}}$. Forward pass is solved for step inputs using z transforms, followed by algebraic modification and subsequent inverse z transform. Eqn. \ref{eqn:set} is solved using Laplace transform. Equivalence under the above-mentioned restriction in sampling interval is established through limiting approximations.
\begin{table*}[t]
	\centering
	\scriptsize
	\caption{Physical model coefficients derived using DiH-RNN for train and test set.}
        \vspace{-0.1in}
	\begin{tabular}{p{1.7 in}|p{0.4 in}|p{0.4 in}|p{0.6 in}|p{0.3 in}|p{0.35 in}|p{0.5 in}|p{0.5 in}|p{0.5 in}}
	 \toprule
		{Train / Test} & {$p_1$ 1/min} & { $p_2$ 1/min} &{$p_3 $ $\frac{10^{-6}}{\mu U.min^2}$} & $p_4$ & {$n$ 1/min} & {$VoI$ dl} & $G_b$ mg/dl & Residue\\ \midrule
Simulation Settings & 0.098 & 0.1406 &	0.028 &	0.05 & 199.6 & -80 & 0.035 & NA \\
Train (mean over all 6 train samples in I2) & 0.0978 & 0.1406 & 0.0262 &	0.0508 & 198.134 & -80.64 & 	0.0349 & 0\\
Test 1 in I1 & 0.0982   & 0.1405  &  0.0256   & 0.0530 & 198.1340  & -80.2774 &    0.0329 & 0.0225\\
Test 2 in I1&     0.0979  &  0.1407 &   0.0274 &   0.0533 & 198.1340 & -85.0589 &    0.0332 & 0.0028\\
Test 3 in I1&     0.0980 &   0.1405 &   0.0262 &   0.0528 & 198.1340 & -85.0973 &    0.0348&0.0011\\
Test 4 in I1&     0.0981  &  0.1405 &   0.0267 &   0.0515 & 198.1340 & -80.6921 &    0.0343 & -0.0168 \\
Test 5 in I1&    0.0979  &  0.1407 &   0.0273 &   0.0548 & 198.1340 & -82.7676 &    0.0317 & 0.0328 \\
Test 6 in I1&     0.0980  &  0.1404 &   0.0275 &   0.0534 & 198.1340 & -82.3447 &    0.0328 & 0.0048\\

		\bottomrule
	\end{tabular}
	\label{tbl:Ex2}
	\vspace{-0.1 in}
\end{table*}
\subsection{Backpropagation to learn coefficients}
The main aim of backpropagation is to derive the approximate coefficient matrices $\mathcal{A}^a$ and $\mathcal{B}^a$. Given an error factor of $\Psi$, we have established that the forward pass is convergent and estimation error is proportional to $\Psi$ if $\tau \leq \frac{\sqrt{2\Psi}}{|a_{ii}|} \forall i$. However, we do not know $a_{ii}$ and hence setting $\tau$ is a difficult task. Often $\tau$ is limited by the sampling frequency of the sensor. In this paper, we assume that the $\tau$ satisfies the condition for convergence of the forward pass. Proposition 1 in \cite{yuan2021physics} shows that for shallow DNNs if all the weights are nonnegative and the activation function is convex and non-decreasing then the overall loss is convex. In a scenario, with only a single minima the gradient descent mechanism is guaranteed to find it.
\section{Artificial Pancreas Example}
\label{sec:Example}

For the AP example, the input set $\Theta$ consists of insulin bolus and meal intake. The set $\Theta$ was constructed by varying bolus value from 0 to 40 U while the meal intake was varied from 0 grams to 28 grams. The model $M(\theta)$ for the AP was the T1D simulator, which is an FDA-approved simulator and widely used for evaluating AP controllers~\cite{visentin2018uva}. The subset of $\theta \in \Theta$ used as sample traces that have no unknown errors given by the following vector:

\begin{scriptsize}
\begin{eqnarray}
\{Bolus, Meal\} = \{(12, 17), (28, 20), (7,6), (14,13), (17, 14),\\\nonumber
(32, 27),  (15,17), (20,20), (10,12), (12,14 ), (25, 22), (5, 12) \}
\end{eqnarray}   
\end{scriptsize}
The PGSM is the BMM discussed in Eqn \ref{eqn:5},\ref{eqn:6}, \ref{eqn:7} with parameter set $\omega$ as shown in Table \ref{tbl:humanEx}. The STL $\phi_\mu$ is globally true that coefficients $\omega$ does not deviate from $\omega_{sim}$, and robustness is given in Eqn. \ref{eqn:phi}.
\begin{equation}
\label{eqn:phi}
\scriptsize
\rho(\phi_\mu,\omega) = max_{i \in \{1 \ldots 7\}}{abs(\frac{\omega[i] - \omega_{sim}[i])}{\omega_{sim}[i]})} - 0.01, 
\end{equation}
where $\omega_{sim}$ is the T1D simulator settings. We partition the input space $\{Bolus,Meal\}$ into test $I_2 = \{(12, 17), (28, 20), (7,6), (14,13), (17, 14), (32, 27)\}$ and train set $I_1 = \{(15,17), (20,20), (10,12), (12,14 ), (25, 22),$ $(5, 12) \}$. The DiH-RNN of AP from Algorithm \ref{alg:Induction} is used to obtain the parameters $\omega$ for the train set as shown in Table \ref{tbl:Ex2}. The residue for each element in the test set is also shown there. Given a probability threshold $1-\alpha = 0.95$ we obtain $d$ to be at position $\lceil (6/2+1)*0.95 \rceil = 4$, i.e. $d = 0.0048$. The interval for the robustness value is $[-0.0216$ $0.0376]$.

Table \ref{tbl:humanEx} shows that for the insulin cartridge problem, the model conformance explained in \ref{sec:ModelConformance} shows that the robustness values under various input configurations are falling outside the range. Hence, these scenarios are deemed to be non-conformal to the original model. Interestingly, all output trajectories satisfied the safety STL $\phi_t$, which is globally true that $CGM > 70 mg/dl$. However, since we evaluate STL robustness on the PGSM coefficients, errors were detected.  

\begin{table*}[t]
	\centering
	\caption{Comparison of physical model coefficients derived using DiH-RNN for different Insulin Blockages, D in the robustness column means error detected and Robustness value from Eqn \ref{eqn:phi} is beyond [-0.0216, 0.0376]. Insulin = 7.5 U, Meal = 20 grams.}
	\begin{tabular}{p{1.0 in}|p{1.0 in }|p{0.4 in}|p{0.4 in}|p{0.6 in}|p{0.3 in}|p{0.35 in}|p{0.5 in}|p{0.5 in}|p{0.5 in}}
	 \toprule
		{Insulin Block Percentage} &{Time until insulin release}& {$p_1$ 1/min} & { $p_2$ 1/min} &{$p_3 $ $\frac{10^{-6}}{\mu U.min^2}$} & $p_4$ & {$n$ 1/min} & {$VoI$ dl} & $G_b$ mg/dl & Robustness\\ \midrule
20 & 150 & 0.065 &0.018 & 0.033 & 0.098 &0.1404 & 268.55 & -51.46 & 0.37 (D)\\
40 & 120 & 0.053 &0.018 & 0.034 & 0.098 &0.1402 & 287.92 & -68.32 & 0.3885 (D)\\
80 & 90 & 0.068 &0.019& 0.034 & 0.098 &0.1401 & 235.25& -58.68 & 0.36 (D)\\
70 & 70 & 0.068 &0.020& 0.033 & 0.098 &0.1400 & 216.14& -48.12 & 0.43 (D)\\
60 & 50 & 0.068 &0.019& 0.034 & 0.098 &0.1405 & 180.48& -69.76 & 0.35 (D) \\
Phantom 20 & 150 & 0.098 & 0.1402 & 0.0194 & 0.058 &	155.89 &	-54.104 & 0.0269 & 0.32 (D)\\
Phantom 40 & 120 & 0.098&	0.1402&	0.0218&	0.0579&	307.06&	-60.73&	0.0339 & 0.5284 (D)\\
Phantom 80 & 90 & 0.098&	0.1401& 0.0217&	0.0503&	143.43&	-64&	0.0344 & 0.27 (D)\\
Phantom 70 & 70 & 0.098&0.139&	0.0229&	0.0655&	169.20&	-48.26&	0.0348&0.48 (D)\\
Phantom 60 & 50 & 0.0983&0.1400&	0.0187&	0.0554&	317.86&	-55.12& 0.0349 & 0.5825 (D)\\
		\bottomrule
	\end{tabular}
	\label{tbl:humanEx}
	\vspace{-0.1 in}
\end{table*}

\section{Conclusions}
In this paper, we propose a methodology to detect unknown unknowns in AI enabled autonomous systems. The principal contribution is integrated modeling of AAS and humans as Human-in-Loop Human-in-Plant (HIL-HIP) systems. We introduce the dynamic-induced hybrid RNN-based (DiH-RNN) learning methodology to derive the model coefficients of the PGSM from operational characteristics. Using these physics-guided models, we utilize statistical model conformance strategy to identify unknown unknowns whose effects remain hidden in the inner parameters of the system and show minimal effect on the outputs. We monitor the operational characteristics of the system to derive any deviation from underlying physical laws. We also showcase the practical applicability of the method in the Artificial Pancreas case study to detect the insulin cartridge problem.
\bibliographystyle{abbrv}
\bibliography{demo,demo1}
	
\end{document}